\documentclass[a4paper,twoside]{article}

\usepackage{epsfig}
\usepackage{subcaption}
\usepackage{calc}
\usepackage{amssymb}
\usepackage[nocompress]{cite}
\usepackage{amstext}
\usepackage{amsmath}
\usepackage{amsthm}
\usepackage{multicol}
\usepackage{pslatex}
\usepackage{apalike}
\usepackage{algorithm}
\usepackage{algpseudocode}
\usepackage{url}
\usepackage[bottom]{footmisc}
\usepackage{svg}
\usepackage{SCITEPRESS}     

\begin{document}

\title{Agent-as-a-Graph: Knowledge Graph-Based Tool and \\ Agent Retrieval for LLM Multi-Agent Systems}

\author{
   \authorname{
       Faheem Nizar,
       Elias Lumer,
       Anmol Gulati, \\
       Pradeep Honaganahalli Basavaraju,
       and Vamse Kumar Subbiah
   }
   \affiliation{Commercial Technology and Innovation Office, PricewaterhouseCoopers, U.S.}
}


\keywords{Large Language Models, Tool Retrieval, Agent Routing, Multi-Agent Systems, Model Context Protocol (MCP)}

\abstract{Recent advances in Large Language Model Multi-Agent Systems enable scalable orchestration and retrieval of specialized, parallelized subagents, each equipped with hundreds or thousands of Model Context Protocol (MCP) servers and tools. However, existing agent, MCP, and retrieval methods typically match queries against a single agent description, obscuring fine-grained tool capabilities of each agent, resulting in suboptimal agent selection. We introduce Agent-as-a-Graph retrieval, a knowledge graph retrieval augmented generation approach that represents both tools and their parent agents as nodes and edges in a knowledge graph. During retrieval, i) relevant agents and tool nodes are first retrieved through vector search, ii) we apply a type-specific weighted reciprocal rank fusion (wRRF) for reranking tools and agents, and iii) parent agents are traversed in the knowledge graph for the final set of agents. We evaluate Agent-as-a-Graph retrieval on the LiveMCPBenchmark, achieving 14.9\% and 14.6\% improvements in Recall@5 and nDCG@5 over prior state-of-the-art retrievers, and 2.4\% improvements in wRRF optimizations.}

\maketitle
\normalsize
\setcounter{footnote}{0}

\section{\uppercase{Introduction}}
\label{sec:introduction}

Recent advances in Large Language Model Multi-Agent Systems enable scalable orchestration and retrieval of specialized, parallelized subagents, each equipped with hundreds or thousands of Model Context Protocol (MCP) servers and tools \cite{qin2023toolllmfacilitatinglargelanguage,li_api_bank_2023,lu2024toolsandboxstatefulconversationalinteractive,langgraph_multi_agent_systems_2024}. With dynamic tool calling and function calling, these agents can execute complex actions such as interacting with data APIs, collaborating on code development, and performing domain-specific question answering \cite{lumer2024toolshedscaletoolequippedagents}. In practice, a single assistant often delegates to specialized subagents for code analysis, databases, or web search, each bundling tens to hundreds of tools behind a single interface. However, with hundreds or even thousands of accessible agent bundles and tool servers, identifying the right agent and the specific tool within it for a given query is increasingly challenging \cite{chen2024reinvoketoolinvocationrewriting,moon_efficient_2024,lumer2025memtooloptimizingshorttermmemory,lumer2025graphragtoolfusion,liu2023controlllmaugmentlanguagemodels}.

Despite advancements in retriever-based tool and agent selection systems, existing agent, MCP, and retrieval methods typically match queries against a single agent description, obscuring fine-grained tool capabilities of each agent and resulting in suboptimal agent selection. Existing strategies typically fall into two camps. Agent-only retrievers match the query against a brief agent description, then operate only within that agent, which can hide an agent's relevant MCP servers and tools when the agent description does not align with the query \cite{mo2025livemcpbenchagentsnavigateocean,du_anytool_2024,huang2024metatoolbenchmarklargelanguage}. Conversely, tool-only retrieval treats each tool independently and ignores the complementary benefits of the surrounding bundle on multi-step tasks \cite{lumer2024toolshedscaletoolequippedagents,moon_efficient_2024,yuan_easytool_2024,chen2024reinvoketoolinvocationrewriting}. Recent benchmarks have highlighted these challenges in multi-step tool selection, evaluating agents across diverse tool repositories \cite{li_api_bank_2023,qin2023toolllmfacilitatinglargelanguage,huang2024metatoolbenchmarklargelanguage}. Furthermore, prior state-of-the-art approaches \cite{mo2025livemcpbenchagentsnavigateocean} have concatenated the agent descriptions and tool descriptions in the same vector space, but the concatenated field suffers from multi-hop query limitations in retrieval augmented generation \cite{peng2024graphretrievalaugmentedgenerationsurvey}.

\begin{figure*}[t]
    \centering
    \includegraphics[width=\linewidth]{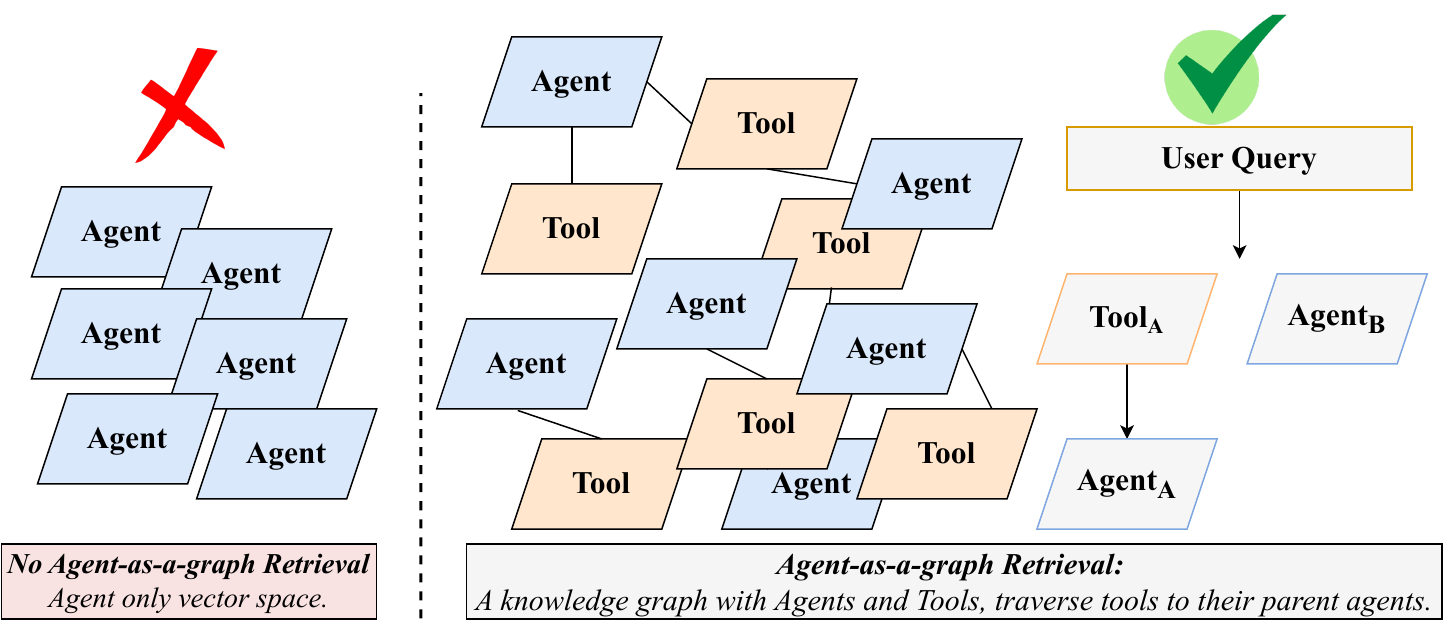}
    \caption{Comparison of agent-only retrieval (left) and the proposed Agent-as-a-Graph (right), which embeds tools and agents as nodes in a knowledge graph with a shared vector space to support joint retrieval and graph traversal.}
    \label{fig:main_diagram}
\end{figure*}

To solve this limitation, we introduce \textit{Agent-as-a-Graph}, a knowledge graph retrieval augmented generation approach that represents both tools and their parent agents as nodes and edges in a knowledge graph. During retrieval, i) relevant agents and tool nodes are first retrieved through vector search, ii) we apply a type-specific weighted reciprocal rank fusion (wRRF) for reranking tools and agents, and iii) parent agents are traversed in the knowledge graph for the final set of agents. We evaluate Agent-as-a-Graph on the LiveMCPBenchmark, achieving 14.9\% and 14.6\% improvements over prior state-of-the-art retrievers in Recall@5 and nDCG@5 respectively, and 2.4\% improvements in wRRF optimizations.

\textbf{\textit{Contributions.}} We make the following contributions through this work.
\begin{enumerate}
    \item \textbf{Agent-as-a-Graph retrieval.} A knowledge graph retrieval augmented generation approach that represents tools and their respective agents as nodes and edges, enabling a unified retrieval approach through graph traversal.
    \item \textbf{Weighted Reciprocal Rank Fusion.} We propose a type-specific weighting mechanism for reciprocal rank fusion (wRRF) after initially retrieving agents and tools, which improves non-reranking approaches by 2.41\%. 
    \item \textbf{Comprehensive Evaluation.} We evaluate Agent-as-a-Graph on the LiveMCPBenchmark across eight embedding models, achieving 14.9\% and 14.6\% improvements in Recall@5 and nDCG@5 over prior state-of-the-art retrievers, and 2.41\% improvements in wRRF optimizations.
\end{enumerate}


\section{\uppercase{Related Works}}
\label{sec:related}

\subsection{Agent Routing and Tool Selection}
Tool selection and agent routing establish the problem of identifying the most relevant tools or agents from large repositories to answer user queries. Early retriever-based methods rely on lexical matching such as TF-IDF and BM25 to align queries with tool descriptions \cite{papineni_why_2001,robertson_probabilistic_2009}. Recent work builds tool knowledge bases and dense retrievers that rank individual tools for a given query, often demonstrating strong single-pass performance at scale \cite{lumer2024toolshedscaletoolequippedagents,lumer2025tooltoagentretrieval,moon_efficient_2024,li_api_bank_2023,yuan_easytool_2024}. Beyond static matching, query and invocation rewriting methods mitigate mismatch between user intent and tool specifications \cite{chen2024reinvoketoolinvocationrewriting}, while progressive and iterative selection decomposes complex tasks to improve recall \cite{anantha_protip_2023,trivedi_interleaving_2023,tang_multihop-rag_2024}. Hierarchical and self-reflective agents decide whether to use tools and which ones, but they typically commit to either an agent-first or a tool-only granularity per decision step \cite{du_anytool_2024,huang2024metatoolbenchmarklargelanguage,langgraph_multi_agent_systems_2024}.

Despite these advances, existing agent, MCP, and retrieval methods typically match queries against a single agent description, obscuring fine-grained tool capabilities of each agent and resulting in suboptimal agent selection \cite{LiveMCPBench_eval_questions_github,lumer2025scalemcpdynamicautosynchronizingmodel}. Agent-first pipelines restrict selection to a single agent scope, potentially missing relevant tools whose parent description does not align with the query, while tool-only retrieval ignores the complementary benefits of coherent agent bundles on multi-step tasks. Recent benchmarks have highlighted these challenges in multi-step tool selection across diverse tool repositories \cite{li_api_bank_2023,qin2023toolllmfacilitatinglargelanguage,huang2024metatoolbenchmarklargelanguage}. In contrast to heavy LLM re-rankers and corrective pipelines \cite{sun_is_2023,yan_corrective_2024}, our approach adopts a transparent, non-learned weighted scoring that combines dense similarity with interpretable type-specific priors to balance tool versus agent emphasis \cite{kuzi2020leveragingsemanticlexicalmatching}. This keeps routing efficient and controllable while avoiding brittle two-stage commitments.

\subsection{Graph-Based Routing and Retrieval}
A complementary line of work explicitly models tools and their relations as graphs to guide selection, planning, or reasoning. ToolNet constructs a tool graph to connect LLMs with massive tool sets \cite{liu2024toolnetconnectinglargelanguage}, while ControlLLM searches over graphs to augment tool use \cite{liu2023controlllmaugmentlanguagemodels}. Prompt-level graph augmentation also appears in Graph-ToolFormer \cite{zhang2023graphtoolformerempowerllmsgraph}. Within Graph RAG, surveys and systems integrate structural signals into retrieval and summarization to improve reasoning, retrieval, question answering, and multi-hop inference \cite{peng2024graphretrievalaugmentedgenerationsurvey,neo4j_graphrag_2024,he2024gretrieverretrievalaugmentedgenerationtextual,edge2024localglobalgraphrag,hu2024graggraphretrievalaugmentedgeneration,sarmah2024hybridragintegratingknowledgegraphs}. Graph-centric reasoning modules such as Think-on-Graph and Graph Chain-of-Thought leverage graph structure for faithful, multi-hop inference \cite{sun2024thinkongraphdeepresponsiblereasoning,jin2024graphchainofthoughtaugmentinglarge}. Additional work explores knowledge graph construction for task planning and API selection but does not evaluate tool retrieval via knowledge graph retrieval augmented generation approaches \cite{Pan_2024,feng2024ontologygroundedautomaticknowledgegraph}.

Unlike these approaches, Agent-as-a-Graph retrieval treats agents and tools as co-equal nodes in a knowledge graph, retrieves over a joint knowledge graph and vector index, and traverses tool-to-agent edges at query time to preserve bundle context. Unlike manual knowledge graph pipelines, we automatically induce the agent-tool knowledge graph from existing MCP manifests and schemas, storing it in standard graph backends and remaining compatible with graph querying languages such as Cypher \cite{neo4j,ozsoy2024text2cypherbridgingnaturallanguage}. Our non-learned type-specific weighting over node types complements graph traversal by deterministically biasing rankings toward agents or tools based on interpretable priors, echoing hybrid lexical-semantic retrieval insights while sidestepping expensive LLM re-ranking \cite{robertson_probabilistic_2009,kuzi2020leveragingsemanticlexicalmatching,sun_is_2023,yan_corrective_2024}. To our knowledge, Agent-as-a-Graph retrieval is the first approach combining knowledge graph structure with type-specific weighted reciprocal rank fusion for agent and tool retrieval in multi-agent systems.


\begin{algorithm}
\footnotesize
\caption{\small Agent-as-a-Graph Retrieval}
\label{alg:aag}
\begin{algorithmic}[1]

\Require query $q$, corpora $\mathcal{C_T},\mathcal{C_A}$, 
weights $(\alpha_{\mathcal{A}},\alpha_{\mathcal{T}})$,
owner map $\mathrm{own}(\cdot)$, cutoffs $N,K$, fusion constant $k$

\State $\mathcal{L_T}\!\gets\!\textsc{TopN}(q,\mathcal{C_T},N)$;\;
       $\mathcal{L_A}\!\gets\!\textsc{TopN}(q,\mathcal{C_A},N)$

\State $\mathcal{L}\!\gets\!\mathcal{L_T}\cup\mathcal{L_A}$

\State \textbf{// type-specific weighted RRF}
\For{$e\in\mathcal{L}$}
    \State $s(e)\gets 
        \begin{cases}
        \alpha_{\mathcal{T}}\!/\!(k+r(e)), & e\in\mathcal{C_T}\\
        \alpha_{\mathcal{A}}\!/\!(k+r(e)), & e\in\mathcal{C_A}
        \end{cases}$
\EndFor

\State Sort $\mathcal{L}$ by $s(e)$ descending

\State \textbf{// tool → agent graph traversal}
\State $\mathcal{A}^\star\gets\emptyset$
\For{$e$ in $\mathcal{L}$}
    \State $a \gets e$ if agent, else $\mathrm{own}(e)$
    \If{$a$ defined and $a\notin\mathcal{A}^\star$}
         \State add $a$; \textbf{if} $|\mathcal{A}^\star|\!=\!K$ \textbf{break}
    \EndIf
\EndFor

\State \Return $\mathcal{A}^\star$

\end{algorithmic}
\end{algorithm}

\section{\uppercase{Method}}

In this section, we present Agent-as-a-Graph retrieval, a knowledge graph-based approach for agent and tool selection in large-scale multi-agent systems. Our method consists of three main components: (1) knowledge graph construction and indexing (\ref{subsec:kg-construction}), (2) knowledge graph retrieval (\ref{subsec:kg-retrieval}), and (3) weighted reciprocal rank fusion for knowledge graph nodes (\ref{subsec:wrrf}). As illustrated in Figure~\ref{fig:main_diagram}, Agent-as-a-Graph embeds both tools and their parent agents in a unified vector space while representing them as nodes in a knowledge graph, explicitly linking each tool to its parent agent through metadata edges. Unlike existing approaches that match queries against a single agent description or treat tools in isolation, our knowledge graph structure preserves fine-grained tool capabilities while maintaining agent-level context through graph traversal.

\subsection{Knowledge Graph Construction and Indexing}\label{subsec:kg-construction}

We consider a catalog of Model Context Protocol (MCP) servers with their corresponding agents, denoted as \(a \in \mathcal{A}\). Each agent \(a\) owns a set of tools \(\mathcal{T}_a\), consisting of API calls, functions, or actions exposed by the agent. The combined system is modeled as a bipartite knowledge graph \(G = (\mathcal{A}, \mathcal{T}, E)\), where edges \(E\) represent ownership relations between tools and agents.

\begin{table*}[t]
\caption{Results on the LiveMCPBench benchmark comparing Agent-as-a-Graph with baselines (BM25, Q.Retrieval, ScaleMCP, and MCPZero). 
Metrics are Recall@K, mAP@K, and nDCG@K for $K \in \{1, 5, 10\}$. Embedding model used is OpenAI text-embedding-ada-002}
\label{tab:results_primary}
\centering
\begingroup
\small 
\setlength{\tabcolsep}{3pt} 
\renewcommand{\arraystretch}{1.05} 
\begin{tabular}{l|ccc|ccc|ccc}
\hline
\textbf{Approach} & \multicolumn{3}{c|}{\textbf{Recall}} & \multicolumn{3}{c|}{\textbf{mAP}} & \multicolumn{3}{c}{\textbf{nDCG}} \\
 & @1 & @3 & @5 & @1 & @3 & @5 & @1 & @3 & @5 \\
\hline
BM25 & 0.205 & 0.20 & 0.20 & 0.12 & 0.12 & 0.12 & 0.14 & 0.14 & 0.14 \\
Q.Retrieval & 0.31 & 0.47 & 0.56 & 0.31 & 0.31 & 0.24 & 0.310 & 0.35 & 0.32 \\
MCPZero & 0.44 & 0.66 & 0.70 & 0.45 & 0.39 & \underline{0.31} & 0.45 & 0.46 & \underline{0.41} \\
ScaleMCP & \underline{0.49} & \underline{0.68} & \underline{0.74} & \underline{0.49} & \underline{0.40} & 0.29 & \underline{0.49} & \underline{0.48} & 0.40 \\
\textbf{Agent-as-a-Graph} & \textbf{0.53} & \textbf{0.80} & \textbf{0.85} & \textbf{0.53} & \textbf{0.46} & \textbf{0.35} & \textbf{0.53} & \textbf{0.55} & \textbf{0.47} \\
\hline
\end{tabular}
\endgroup
\end{table*}

\subsubsection{Bipartite Knowledge Graph Modeling}

This knowledge graph representation directly addresses the limitations of single-level retrieval that obscure either tool-level specificity or agent-level context. Retrieving solely by agent descriptions can miss fine-grained functional capabilities at the tool level, while retrieving tools independently discards important execution context such as authentication, parameter inference, or access policies maintained at the agent level. By integrating both levels in a knowledge graph structure, the retriever can surface relevant tools without losing their surrounding agent context through graph traversal. Furthermore, every query or sub-query ultimately resolves to an executable agent through the graph edges, ensuring that the retrieved entity can act upon the user request.

\subsubsection{Node Representation and Indexing}

We construct a unified agent-tool knowledge graph with associated catalog \(\mathcal{C}\) that integrates both tools and agents for retrieval. The catalog is composed of two node types: the \emph{tool corpus} \(\mathcal{C_T}\) and the \emph{agent corpus} \(\mathcal{C_A}\).

\textbf{Tool Nodes} \(\mathcal{C_T} \subset \mathcal{C}\) contain tool names and descriptions that are directly indexed for retrieval in the knowledge graph. Each tool node includes metadata edges explicitly linking it to its parent MCP server or agent node, denoted as \(owner(\mathcal{T}) = \mathcal{A}\). This edge structure enables graph traversal from a retrieved tool to the corresponding executable agent during query resolution.

\textbf{Agent Nodes} \(\mathcal{C_A} \subset \mathcal{C}\) similarly contain agent names and descriptions, representing higher-level capabilities and serving as parent nodes within the knowledge graph structure. Agent nodes aggregate the capabilities of their associated tools, providing coarser-grained retrieval targets when tool-level specificity is not required.

\subsection{Agent-as-a-Graph Retrieval}\label{subsec:kg-retrieval}

The retrieval process modifies the standard top-\(K\) ranking procedure to operate over our knowledge graph structure and associated catalog. The objective is to identify the top-\(K\) most relevant agents for a given query or sub-query. To achieve this, we first retrieve the top \(N \gg K\) nodes from the knowledge graph catalog \(\mathcal{C}\), ranked by semantic similarity to the query. The corresponding parent agents are then aggregated through graph traversal along the ownership edges, and the top-\(K\) unique agents are selected. The complete knowledge graph retrieval procedure is detailed in Algorithm~\ref{alg:aag}.

\subsubsection{Query Processing}

The input to Agent-as-a-Graph can be the original user query, the decomposed sub-steps derived from it, or a combination of both. We evaluate two query paradigms.

\textbf{Direct Querying} uses the user's high-level question directly as the retrieval query without any pre-processing. This approach retrieves the top-\(K\) most relevant agents or tools for the overall task.

\textbf{Step-wise Querying} decomposes the original query into a sequence of smaller sub-tasks. Each step is then submitted independently to the retriever, allowing the system to identify different agents as needed across multi-step workflows. This decomposition strategy aligns with reasoning-based retrieval planning methods that break complex queries into manageable sub-goals. This step-wise procedure is the primary setting used in our evaluations.

\begin{table*}[t]
\centering
\caption{Per-embedding model comparison of Agent-as-a-Graph against MCPZero}
\label{tab:retriever_comparison}
\begingroup
\small
\setlength{\tabcolsep}{4pt}
\renewcommand{\arraystretch}{1.05}
\begin{tabular}{l|cc|cc|cc}
\hline
\textbf{Retriever Model} & 
\multicolumn{2}{c|}{\textbf{Recall@5}} & 
\multicolumn{2}{c|}{\textbf{nDCG@5}} & 
\multicolumn{2}{c}{\textbf{mAP@5}} \\ 
\cline{2-7}
 & Ours & MCPZero & Ours & MCPZero & Ours & MCPZero \\
\hline
\footnotesize{Vertex AI text-embedding-005} & \textbf{0.87} & 0.74 & \textbf{0.48} & 0.42 & \textbf{0.36} & 0.32 \\
\footnotesize{Gemini-embedding-001} & \textbf{0.86} & 0.74 & \textbf{0.49} & 0.44 & \textbf{0.37} & 0.34 \\
\footnotesize{Amazon titan-embed-text-v2} & \textbf{0.85} & 0.66 & \textbf{0.47} & 0.37 & \textbf{0.35} & 0.28 \\
\footnotesize{Amazon titan-embed-text-v1} & \textbf{0.85} & 0.65 & \textbf{0.48} & 0.39 & \textbf{0.36} & 0.30 \\
\footnotesize{OpenAI text-embedding-ada-002} & \textbf{0.83} & 0.70 & \textbf{0.50} & 0.40 & \textbf{0.39} & 0.30 \\
\footnotesize{OpenAI text-embedding-3-small} & \textbf{0.87} & 0.72 & \textbf{0.49} & 0.41 & \textbf{0.36} & 0.31 \\
\footnotesize{OpenAI text-embedding-3-large} & \textbf{0.87} & 0.74 & \textbf{0.50} & 0.42 & \textbf{0.38} & 0.32 \\
\footnotesize{All-MiniLM-L6-v2} & \textbf{0.80} & 0.67 & \textbf{0.45} & 0.39 & \textbf{0.33} & 0.29 \\
\hline
\end{tabular}
\endgroup
\end{table*}

\subsection{Weighted Reciprocal Rank Fusion for Knowledge Graph Nodes}\label{subsec:wrrf}

Our approach combines retrieval from fundamentally different node types in the knowledge graph (\(\mathcal{C_T}\) and \(\mathcal{C_A}\)). Assigning per-type weights lets us bias ranking toward the granularity that best answers the query while preserving bundle context during graph traversal to the executable agent. Empirically, modest emphasis on agent nodes improves top-\(K\) routing stability while retaining enough sensitivity to surface highly specific tool nodes when queries are fine-grained (\ref{sec:evaluation}).

\subsubsection{Standard Reciprocal Rank Fusion}

Standard Reciprocal Rank Fusion (RRF) \cite{36196} combines multiple ranked lists by summing the reciprocal of each entity's rank with a constant \(k\) to dampen outlier systems:
\begin{equation}
 s_{RRF}(e) = \sum_{m \in \mathcal{M}} \frac{1}{k + r_m(e)}    
\end{equation}
 
Here, the different methods \(m \in \mathcal{M}\) correspond to retrieved elements from \(\mathcal{C_T}\) and \(\mathcal{C_A}\). RRF is attractive because it is training-free, robust across heterogeneous signals, and consistently competitive against stronger baselines. A practical generalization considers different weights for different retrieval methods \(m \in \mathcal{M}\), where each weight is given as \(\alpha_m\):
\begin{equation}\label{eq:wRRF}
    s_{wRRF}(e) = \sum_{m \in \mathcal{M}} \frac{\alpha_m}{k + r_m(e)}
\end{equation}

\subsubsection{Type-Specific Weighted RRF}

For our retrieval method, a key design choice is to decouple tools and agents at scoring time. We first form a single unified candidate list from the merged corpus \(\mathcal{C}=\mathcal{C_A}\cup\mathcal{C_T}\) and assign each entity a global base rank \(r(e)\), independent of any specific retriever after consolidation. We then adapt weighted Reciprocal Rank Fusion (wRRF) by applying type-specific weights. For a small damping constant \(k\) (commonly \(k=60\)) and entity \(e\in\mathcal{C}\), the type-conditioned score is:
\begin{equation} \label{eq:ourRRF}
    s_{new}(e)=
        \begin{cases}
        \displaystyle \frac{\alpha_{\mathcal{T}}}{\,k + r(e)\,}, & e \in \mathcal{C_T},\\[8pt]
        \displaystyle \frac{\alpha_{\mathcal{A}}}{\,k + r(e)\,}, & e \in \mathcal{C_A}.
    \end{cases}
\end{equation}

We rerank all candidates by \(s_{\text{new}}(e)\) and select the top-\(K\) with ties broken by smaller \(r(e)\). Decoupling tools and agents at scoring time avoids blending heterogeneous calibration from distinct retrievers while retaining the calibration-insensitive strengths of RRF. 

The piecewise form of \(s_{new}(e)\) provides three key benefits. First, it offers interpretability via two clear knobs \(\alpha_{\mathcal{A}}\) and \(\alpha_{\mathcal{T}}\) that allow explicit control over agent versus tool emphasis. Second, it ensures stability in top-\(K\) routing by maintaining consistent relative rankings within each node type. Third, it provides flexibility to shift the balance between agent coverage and tool specificity based on query characteristics. Empirically, this led to higher Recall@\(K\) and nDCG@\(K\) in our benchmarks, particularly for mixed-intent workloads where both agent orchestration and tool specificity are relevant (\ref{sec:evaluation}).



\section{Evaluations}
\label{sec:evaluation}

We evaluate Agent-as-a-Graph against classical and contemporary baseline methods across three experiments: baseline comparison, cross-architecture generalization, and type-weighted ranking. Our evaluation demonstrates that knowledge graph-based agent-tool indexing substantially improves retrieval accuracy, generalizes across embedding architectures, and benefits from type-specific weighting.

\subsection{Experimental Setup}

\textbf{Dataset.} We evaluate on the LiveMCPBench benchmark \cite{mo2025livemcpbenchagentsnavigateocean,LiveMCPBench_eval_questions_github}, which contains 70 MCP servers and 527 tools, with 95 real-world questions annotated with step-by-step breakdowns and relevant agent-tool mappings. On average, each question spans 2.68 steps and involves 2.82 tools and 1.40 MCP agents. To isolate the contribution of tool-level information, an agent-only baseline dataset is constructed containing only MCP server names and descriptions.

\begin{table*}[t]
\caption{Comparison of weighted RRF approaches with different type-specific weight ratios \((\alpha_{\mathcal{A}}, \alpha_{\mathcal{T}})\). Our method (Equation~\ref{eq:ourRRF}) outperforms standard weighted RRF at balanced configurations. Embedding model used is OpenAI text-embedding-ada-002}
\label{tab:results_weighted}
\centering
\begingroup
\small
\setlength{\tabcolsep}{3pt}
\renewcommand{\arraystretch}{1.05} 
\begin{tabular}{l|cc|cc|cc}
\hline
\((\alpha_{\mathcal{A}}, \alpha_{\mathcal{T}})\) & \multicolumn{2}{c|}{Recall@5} & \multicolumn{2}{c|}{mAP@5} & \multicolumn{2}{c}{nDCG@5} \\
 & Ours & RRF & Ours & RRF & Ours & RRF\\
\hline
(1.0, 3.0) & \textbf{0.80} & 0.79 & 0.33 & \textbf{0.34} & 0.44 & \textbf{0.45} \\
(1.0, 2.0) & 0.80          & 0.80 & 0.33 & \textbf{0.33} & 0.44 & \textbf{0.45} \\
(1.0, 1.5) & 0.80          & 0.80 & 0.33 & \textbf{0.34} & 0.44 & \textbf{0.45} \\
(1.0, 1.0) & \textbf{0.83} & 0.79 & \textbf{0.34} & 0.33 & \textbf{0.46} & 0.44 \\
(1.5, 1.0) & \textbf{0.85} & 0.79 & \textbf{0.35} & 0.33 & \textbf{0.47} & 0.44 \\
(2.0, 1.0) & \textbf{0.82} & 0.80 & 0.33 & \textbf{0.34} & 0.45 & \textbf{0.45} \\
(3.0, 1.0) & 0.76 & \textbf{0.80} & 0.32 & \textbf{0.34} & 0.43 & \textbf{0.45} \\
\hline
\end{tabular}
\endgroup
\end{table*}

\textbf{Baselines.} Agent-as-a-Graph is compared against BM25 \cite{llamaindex_bm25_retriever} for lexical matching, standard RRF \cite{10.1145/1571941.1572114} for list fusion, and state-of-the-art agent retrievers MCPZero \cite{fei2025mcpzeroactivetooldiscovery} and ScaleMCP \cite{lumer2025scalemcpdynamicautosynchronizingmodel}.

\textbf{Embedding Models.} Retrieval performance is evaluated across eight embedding models: Vertex AI (text-embedding-005, Gemini-embedding-001) \cite{vertex_embedding_models}, Amazon Titan (v1, v2) \cite{titan_embedding_models}, OpenAI (ada-002, 3-small, 3-large) \cite{openai_embedding_models}, and All-MiniLM-L6-v2.

\textbf{Metrics.} Evaluation metrics include Recall@\(K\), mean average precision (mAP@\(K\)), and normalized discounted cumulative gain (nDCG@\(K\)) at \(K \in \{1, 3, 5\}\). Retrieval accuracy compares retrieved agents against ground truth agents for each query.

\textbf{Protocol.} Step-wise querying is employed where each step is submitted independently to the retriever. Semantic similarity search retrieves the top \(N \gg K\) nodes from the knowledge graph catalog, then selects the top-\(K\) unique agents using Algorithm~\ref{alg:aag} through graph traversal. For type-weighting experiments, \(\alpha_{\mathcal{A}}\) and \(\alpha_{\mathcal{T}}\) are swept over a ratio grid including the unweighted point \((1,1)\).

\subsection{Baseline Comparison Experiments}

We compare Agent-as-a-Graph against BM25, q-retrieval, standard RRF, MCPZero, and ScaleMCP on the LiveMCPBench benchmark. For this experiment, we use the unified corpus \(\mathcal{C}=\mathcal{C_A}\cup\mathcal{C_T}\) without type-specific weighting, following Algorithm~\ref{alg:aag}. We use OpenAI text-embedding-ada-002 as the embedding model to obtain similarity scores and retrieve the initial top-\(K\) entities.

\subsubsection{Results}

Table~\ref{tab:results_primary} summarizes performance across Recall@\(K\), mAP@\(K\), and nDCG@\(K\). Agent-as-a-Graph consistently outperforms all baselines across all metrics. Our approach achieves Recall@5 of 0.83 compared to RRF's 0.79, representing an improvement of 5.1\%. Compared to MCPZero's 0.70, Agent-as-a-Graph shows an improvement of 18.6\%. For nDCG@5, Agent-as-a-Graph obtains 0.46 compared to RRF's 0.44 and MCPZero's 0.41. Analysis reveals that 39.13\% of retrieved top-\(K\) items originate from agent nodes \(\mathcal{C_A}\), while 34.44\% of matched tool nodes trace back to agent nodes through graph edges, indicating that both node types contribute meaningfully to retrieval performance.

\subsubsection{Discussion}

These improvements arise primarily from the knowledge graph structure that jointly indexes tools and agents as co-equal nodes, enabling finer-grained semantic alignment while preserving agent context through graph traversal. Importantly, the performance lift is not attributable to tool-level retrieval alone; the fact that over 39\% of retrieved items come from agent nodes indicates that agent-side evidence remains crucial during selection. The knowledge graph structure mitigates context dilution and improves multi-step routing without sacrificing fine-grained precision, validating our hypothesis that unified representation of tools and agents provides a richer search space than agent-only or tool-only approaches.

\subsection{Cross-Architecture Generalization Experiments}

We evaluate Agent-as-a-Graph across eight embedding models to verify architecture-agnostic gains. For each encoder family, we compare MCPZero against Agent-as-a-Graph where for the ranking, we use an unweighted approach, where both tool and agent weights, \(\alpha_{\mathcal{T}}\) and \(\alpha_{\mathcal{A}}\) are equal.

\begin{figure*}[t]
    \centering
    \includegraphics[width=14cm]{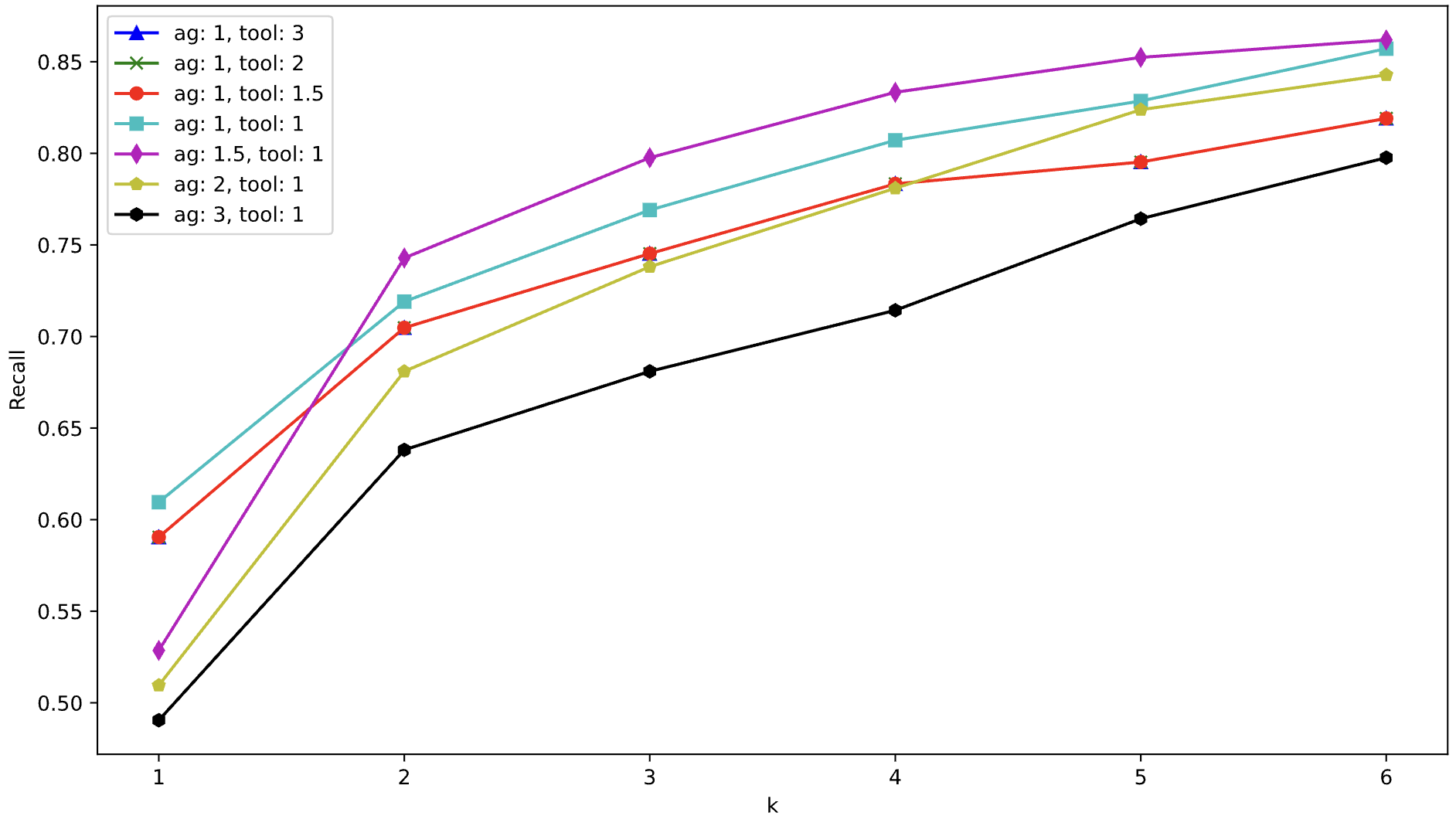}
    \caption{Recall@5 performance across different type-specific weight configurations \((\alpha_{\mathcal{A}}, \alpha_{\mathcal{T}})\). The plot shows that modest agent emphasis (1.5:1) achieves optimal performance while extreme bias (3:1) degrades precision. Agent-as-a-Graph consistently outperforms standard weighted RRF across all configurations.}
    \label{fig:weighted_rrf_recall}
\end{figure*}

\subsubsection{Results}

Table~\ref{tab:retriever_comparison} presents per-embedding comparisons against MCPZero. Agent-as-a-Graph exhibits stable improvements across all encoder families, with remarkably low variance (std.\ \(\approx 0.02\) in Recall@5 and \(0.01\) in nDCG@5). The strongest improvement occurs with Amazon Titan v2, where Recall@5 increases from 0.66 to 0.85, representing an improvement of 28.8\%. Even the compact All-MiniLM-L6-v2 model shows an improvement of 19.4\% in Recall@5, increasing from 0.67 to 0.80, confirming generalizability beyond large proprietary embeddings. On average, Agent-as-a-Graph achieves Recall@5 of 0.85 compared to 0.70 for MCPZero, representing an improvement of 19.4\%. Average nDCG@5 increases from 0.41 to 0.48, an improvement of 17.7\%, while average mAP@5 increases from 0.31 to 0.36. These results demonstrate that Agent-as-a-Graph consistently outperforms MCPZero across model architectures and retrieval metrics.

\subsubsection{Discussion}

The low variance across embedding families (std.\ \(\approx 0.02\)) indicates that gains stem primarily from the structural design of the knowledge graph rather than encoder-specific characteristics. Even compact open-source models like All-MiniLM-L6-v2 achieve substantial improvements, confirming that the method does not require large proprietary embeddings to be effective. The consistency across diverse embedding architectures (from Google's Vertex AI to Amazon's Titan to OpenAI's models) demonstrates that the knowledge graph structure provides fundamental advantages in representing agent-tool relationships that translate across different vector spaces. This robustness is particularly valuable for deployment scenarios where embedding choice may be constrained by cost, latency, or licensing considerations.

Table~\ref{tab:results_weighted} and Figure~\ref{fig:weighted_rrf_recall} present the effects of type-specific weighting on retrieval performance. The optimal configuration (\(\alpha_{\mathcal{A}}=1.5, \alpha_{\mathcal{T}}=1.0\)) achieves Recall@5 of 0.85 and nDCG@5 of 0.47, outperforming the unweighted baseline (Recall@5 of 0.83, nDCG@5 of 0.46) and standard weighted RRF (Recall@5 of 0.79, nDCG@5 of 0.44). 

\subsection{Type-Weighted Ranking Experiments}

We evaluate the effect of type-specific weighting on retrieval accuracy by comparing standard weighted RRF (Equation~\ref{eq:wRRF}) against our type-conditioned approach (Equation~\ref{eq:ourRRF}). We fix \(k=60\) and sweep \(\alpha_{\mathcal{A}}\) and \(\alpha_{\mathcal{T}}\) over a ratio grid including the unweighted point \((1,1)\). We use OpenAI text-embedding-ada-002 as the embedding model to obtain similarity scores and retrieve the initial top-\(K\) entities.

\subsubsection{Results}

Type-specific weights act as transparent knobs that shift emphasis between agent coverage and tool specificity. Overly aggressive agent bias (3:1) reduces Recall@5 to 0.76, while tool-dominant settings (1:3) achieve 0.80, demonstrating the importance of balanced weighting. The performance curve peaks at modest agent emphasis (1.5:1), confirming that slight prioritization of agent nodes benefits routing stability without sacrificing tool-level precision.

\subsubsection{Discussion}

Type-conditioned weights (Equation~\ref{eq:ourRRF}) outperform standard weighted RRF (Equation~\ref{eq:wRRF}) by decoupling node types at scoring time while preserving within-type ordering through the shared denominator \(k+r(e)\). This design provides interpretable control over retrieval granularity without the complexity of learned re-ranking models. The optimal configuration shows modest agent emphasis (1.5:1), balancing routing stability with sensitivity to fine-grained tool matches. The performance degradation at extreme ratios (3:1 agent bias, 1:3 tool bias) confirms that overly aggressive weighting can harm precision. The transparent, tunable nature of type-specific weights makes them particularly valuable for production systems where interpretability and control are priorities. Unlike black-box re-rankers, the weights expose simple knobs \((\alpha_{\mathcal{A}}, \alpha_{\mathcal{T}})\) that can be adjusted based on domain characteristics or query intent, enabling query-adaptive tuning without retraining.

\section{\uppercase{Conclusion}}
\label{sec:conclusion}

As multi-agent systems scale to hundreds of specialized agents coordinating thousands of tools and MCP servers, routing queries to the correct agent remains a fundamental challenge. Traditional approaches that match queries against single agent descriptions obscure the fine-grained capabilities of individual tools, while tool-only retrieval discards valuable agent-level context for multi-step workflows. We presented Agent-as-a-Graph retrieval, a knowledge graph-based framework that represents tools and agents as co-equal nodes linked through explicit metadata edges, enabling unified retrieval through graph traversal and type-specific weighted reciprocal rank fusion. Experiments across eight embedding models on LiveMCPBench demonstrate that Agent-as-a-Graph consistently outperforms classical and contemporary baselines, achieving Recall@5 of 0.85 and nDCG@5 of 0.48 with optimal type-specific weighting, representing improvements of 14.9\% and 14.6\% over prior state-of-the-art retrievers. Analysis reveals that 39.13\% of retrieved items originate from agent nodes while 34.44\% trace from tool nodes through graph edges, confirming that both node types contribute meaningfully to routing performance. Agent-as-a-Graph establishes a foundation for knowledge graph-based retrieval in increasingly complex agent ecosystems and motivates future research into query-adaptive type weights and structure-aware graph priors for multi-agent orchestration.

\clearpage

\bibliographystyle{apalike}
{\small
\bibliography{references}}

\end{document}